\documentclass[preprint,5p,12pt]{article}

\usepackage{amssymb}
\usepackage[position=top]{subfig}
\usepackage{algorithm,algorithmic}
\usepackage{graphicx}
\usepackage{xspace}
\usepackage{multirow}
\usepackage{rotating}
 \usepackage{url}
\usepackage{gensymb}
\usepackage [ table ]{ xcolor }
\usepackage[top=2cm, bottom=2.2cm, left=1.5cm, right=1.5cm]{geometry}
\usepackage{setspace}
\usepackage{morefloats}
\usepackage[keeplastbox]{flushend}

\usepackage{multirow}
\usepackage{caption}
\usepackage{url}
\usepackage[hyphenbreaks]{breakurl}
\usepackage{amsmath}

\usepackage[affil-it]{authblk}
\usepackage[hidelinks]{hyperref}
\hypersetup{
	colorlinks = true,
	citecolor = blue,
}
\usepackage{pdfpages}

\usepackage{newfloat}
	\DeclareFloatingEnvironment[
	fileext=los,
	listname={List of Listings},
	name=Listing,
	placement=tbhp,
]{code}

\usepackage{pdfpages}
\usepackage{eso-pic}
\definecolor{backcream}{HTML}{FFF0C8}

\AddToShipoutPictureFG*{
	\AtPageUpperLeft{%
		\raisebox{-50pt}{\makebox[\paperwidth]{\noindent\fcolorbox{red}{backcream}{\begin{minipage}{18.5cm}\centering
						{\color{red}\textbf{This manuscript is accepted in Frontiers in Neuroscience.} \textbf{Please cite it as:}}\\
						Mozafari, M., Ganjtabesh, M., Nowzari-Dalini, A., \& Masquelier, T. (2019). SpykeTorch: Efficient Simulation of Convolutional Spiking Neural Networks With at Most One Spike per Neuron. Frontiers in Neuroscience (\url{https://doi.org/10.3389/fnins.2019.00625}).
		\end{minipage}}}}%
	}
}

\begin{document}
\title{\LARGE \textbf{SpykeTorch: Efficient Simulation of Convolutional Spiking Neural Networks with at most one Spike per Neuron}}

\author{Milad~Mozafari$ ^{1,2}$}
\author{Mohammad~Ganjtabesh$ ^{1,}$\footnote{Corresponding author. \\ \hspace*{0.5cm} Email addresses:\\ \hspace*{1cm} milad.mozafari@ut.ac.ir (MM), \\ \hspace*{1cm} mgtabesh@ut.ac.ir (MG) \\ \hspace*{1cm}  nowzari@ut.ac.ir (AND) \\ \hspace*{1cm} timothee.masquelier@cnrs.fr (TM).}}
\author{Abbas~Nowzari-Dalini$ ^{1}$}
\author{Timoth\'ee~Masquelier$ ^{2}$}

\affil{\footnotesize $ ^{1} $ Department of Computer Science, School of Mathematics, Statistics, and Computer Science, University of Tehran, Tehran, Iran}
\affil{\footnotesize $ ^{2} $ CerCo UMR 5549, CNRS -- Universit\'e Toulouse 3, France}

\date{}

\maketitle
\begin{abstract}
	Application of deep convolutional spiking neural networks (SNNs) to artificial intelligence (AI) tasks has recently gained a lot of interest since SNNs are hardware-friendly and energy-efficient. Unlike the non-spiking counterparts, most of the existing SNN simulation frameworks are not practically efficient enough for large-scale AI tasks. In this paper, we introduce SpykeTorch, an open-source high-speed simulation framework based on PyTorch. This framework simulates convolutional SNNs with at most one spike per neuron and the rank-order encoding scheme. In terms of learning rules, both spike-timing-dependent plasticity (STDP) and reward-modulated STDP (R-STDP) are implemented, but other rules could be implemented easily. Apart from the aforementioned properties, SpykeTorch is highly generic and capable of reproducing the results of various studies. Computations in the proposed framework are tensor-based and totally done by PyTorch functions, which in turn brings the ability of just-in-time optimization for running on CPUs, GPUs, or Multi-GPU platforms.
	
	\vspace{0.3cm}
\end{abstract}

\section{Introduction} \label{intro}
For many years, scientist were trying to bring human-like vision into machines and artificial intelligence (AI). In recent years, with advanced techniques based on deep convolutional neural networks (DCNNs)~\cite{rawat2017deep,gu2018recent}, artificial vision has never been closer to human vision. Although DCNNs have shown outstanding results in many AI fields, they suffer from being data- and energy-hungry. Energy consumption is of vital importance when it comes to hardware implementation for solving real-world problems.

Our brain consumes much less energy than DCNNs, about $20$ Watts~\cite{mink1981ratio} – roughly the power consumption of an average laptop, for its top-notch intelligence. This feature has convinced researchers to start working on computational models of human cortex for AI purposes. Spiking neural networks (SNNs) are the next generation of neural networks, in which neurons communicate through binary signals known as spikes. SNNs are energy-efficient for hardware implementation, because, spikes bring the opportunity of using event-based hardware as well as simple energy-efficient accumulators instead of complex energy-hungry multiply-accumulators that are usually employed in DCNN hardware~\cite{furber2016large, davies2018loihi}.

Spatio-temporal capacity of SNNs makes them potentially stronger than DCNNs, however, harnessing their ultimate power is not straightforward. Various types of SNNs have been proposed for vision tasks which can be categorized based on their specifications such as:
\begin{itemize}
	\item network structure: shallow~\cite{masquelier2007unsupervised, yu2013rapid,kheradpisheh2016bio}, and deep~\cite{kheradpisheh2018stdp,mozafari2019bioinspired},
	\item topology of connections: convolutional~\cite{cao2015spiking, tavanaei2016bio}, and fully connected~\cite{diehl2015unsupervised},
	\item information coding: rate~\cite{hussain2014improved,oconnor2014realtime}, and latency~\cite{masquelier2007unsupervised, diehl2015unsupervised,mostafa2018supervised},
	\item learning rule: unsupervised~\cite{diehl2015unsupervised,ferre2018unsupervised,thiele2018event}, supervised~\cite{diehl2015fast,wu2018spatio,liu2017mtspike,mostafa2018supervised,shrestha2018slayer,zenke2018superspike,bellec2018long}, and reinforcement~\cite{florian2007reinforcement,mozafari2018first}.
\end{itemize}
For recent advances in deep learning with SNNs, we refer the readers to~\cite{tavanaei2018deep, pfeiffer2018deep, neftci2019surrogate}. 

Deep convolutional SNNs (DCSNNs) with time-to-first-spike information coding and STDP-based learning rule constitute one of those many types of SNNs that carry interesting features. Their deep convolutional structure supports visual cortex and let them extract features hierarchically from simple to complex. Information coding using the earliest spike time, which is proposed based on the rapid visual processing in the brain~\cite{thorpe1996speed}, needs only a single spike, making them super fast and more energy efficient. These features together with hardware-friendliness of STDP, turn this type of SNNs into the best option for hardware implementation and online on-chip training~\cite{yousefzadeh2017hardware}. Several recent studies have shown the excellence of this type of SNNs in visual object recognition~\cite{mozafari2018first, kheradpisheh2018stdp,mostafa2018supervised,mozafari2019bioinspired, falez2019multi, vaila2019deep}.

With simulation frameworks such as Tensorflow~\cite{abadi2016tensorflow} and PyTorch~\cite{paszke2017automatic}, developing and running DCNNs is fast and efficient. Conversely, DCSNNs suffer from the lack of such frameworks. Existing state-of-the-art SNN simulators have been mostly developed for studying neuronal dynamics and brain functionalities and are not efficient and user-friendly enough for AI purposes. For instance, bio-realistic and detailed SNN simulations are provided by NEST~\cite{gewaltig2007nest}, BRIAN~\cite{stimberg2014equation}, NEURON~\cite{carnevale2006neuron}, and ANNarchy~\cite{vitay2015annarchy}. These frameworks also enable users to define their own dynamics of neurons and connections. In contrast, frameworks such as Nengo~\cite{bekolay2014nengo} and NeuCube~\cite{kasabov2014neucube} offer high-level simulations focusing on the neural behavior of the network. Recently, BindsNet~\cite{hazan2018bindsnet} framework has been proposed as a fast and general SNN simulator based on PyTorch that is mainly developed for conducting AI experiments. A detailed comparison between BindsNet and the other available frameworks can be found in their paper.

In this paper, we propose SpykeTorch, a simulation framework based on PyTorch which is optimized specifically for convolutional SNNs with at most one spike per neuron. SpykeTorch offers utilities for building hierarchical feedforward SNNs with deep or shallow structures and learning rules such as STDP and R-STDP~\cite{gerstner1996neuronal, bi1998synaptic,fremaux2016neuromodulated, brzosko2017sequential}. SpykeTorch only supports time-to-first-spike information coding and provides a non-leaky integrate and fire neuron model with at most one spike per stimulus. Unlike BindsNet which is flexible and general, the proposed framework is highly restricted to and optimized for this type of SNNs. Although BindsNet is based on PyTorch, its network design language is different. In contrast, SpykeTorch is fully compatible and integrated with PyTorch and obeys the same design language. Therefore, a PyTorch user may only read the documentation to find out the new functionalities. Besides, this integrity makes it possible to utilize almost all of the PyTorch's functionalities either running on a CPU, or (multi-) GPU platform.

The rest of this paper is organized as follows: Section \ref{time} describes how SpykeTorch includes the concept of time in its computations. Section \ref{pkg} is dedicated to SpykeTorch package structure and its components. In Section \ref{tutor}, a brief tutorial on building, training, and evaluating a DCSNN using SpykeTorch is given. Section \ref{conclusion} summarizes the current work and highlights possible future works. 

\section{Time Dimension}\label{time}
Modules in SpykeTorch are compatible with those in PyTorch and the underlying data-type is simply the PyTorch's tensors. However, since the simulation of SNNs needs the concept of ``time", SpykeTorch considers an extra dimension in tensors for representing time. The user may not need to think about this new dimensionality while using SpykeTorch, but, in order to combine it with other PyTorch's functionalities or extracting different kinds of information from SNNs,
it is important to have a good grasp of how SpykeTorch deals with time.

SpykeTorch works with time-steps instead of exact time. Since the neurons emit at most one spike per stimulus, it is enough to keep track of the first spike times (in time-step scale) of the neurons. For a particular stimulus, SpykeTorch divides all of the spikes into a pre-defined number of spike bins, where each bin corresponds to a single time-step. More precisely, assume a stimulus is represented by $F$ feature maps, each constitutes a grid of $H \times W$ neurons. Let $T_{max}$ be the maximum possible number of time-steps (or bins) and $T_{f,r,c}$ denote the spike time (or the bin index) of the neuron placed at position $(r,c)$ of the feature map $f$, where $0\leq f < F$, $0\leq r < H$, $0\leq c < W$, and $T_{f,r,c} \in \{0,1, ..., T_{max}-1\} \cup \{\infty\}$. The $\infty$ symbol stands for no spike. SpykeTorch considers this stimulus as a four-dimensional binary spike-wave  tensor $S$ of size $T_{max} \times F \times H \times W$ where:
\begin{equation}
S[t,f,r,c] = \begin{cases}
0 & t < T_{f,r,c},\\
1 & \textit{otherwise}.
\end{cases}
\end{equation}

\begin{figure}
	\begin{center}
		\includegraphics[width=0.6\textwidth]{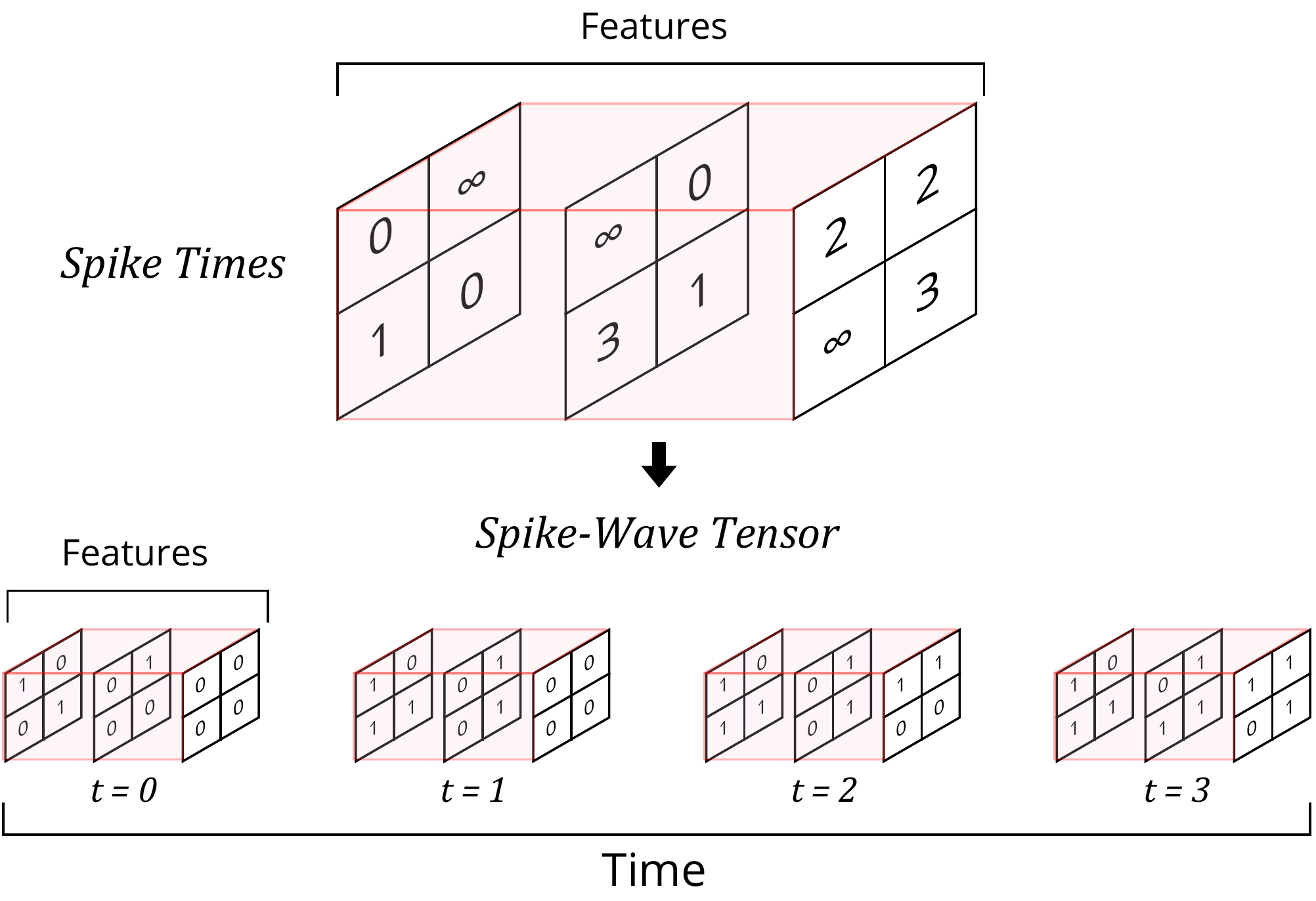}
	\end{center}
	\caption{An example of generating spike-wave tensor from spike times. There are three feature maps, each constitutes a $2\times 2$ grid of neurons that are going to generate a spike-wave representing a stimulus. If the maximum number of time-steps is $4$, then the resulting spike-wave is a four-dimensional tensor of size $4\times 3\times 2\times 2$. If a neuron emits spike at time step $t = i$, the corresponding position in the spike-wave tensor will be set to $1$ from time-step $t = i$ to the final time step ($t = 3$).}\label{fig:1}
\end{figure}

\begin{figure}
	\begin{center}
		\includegraphics[width=0.5\textwidth]{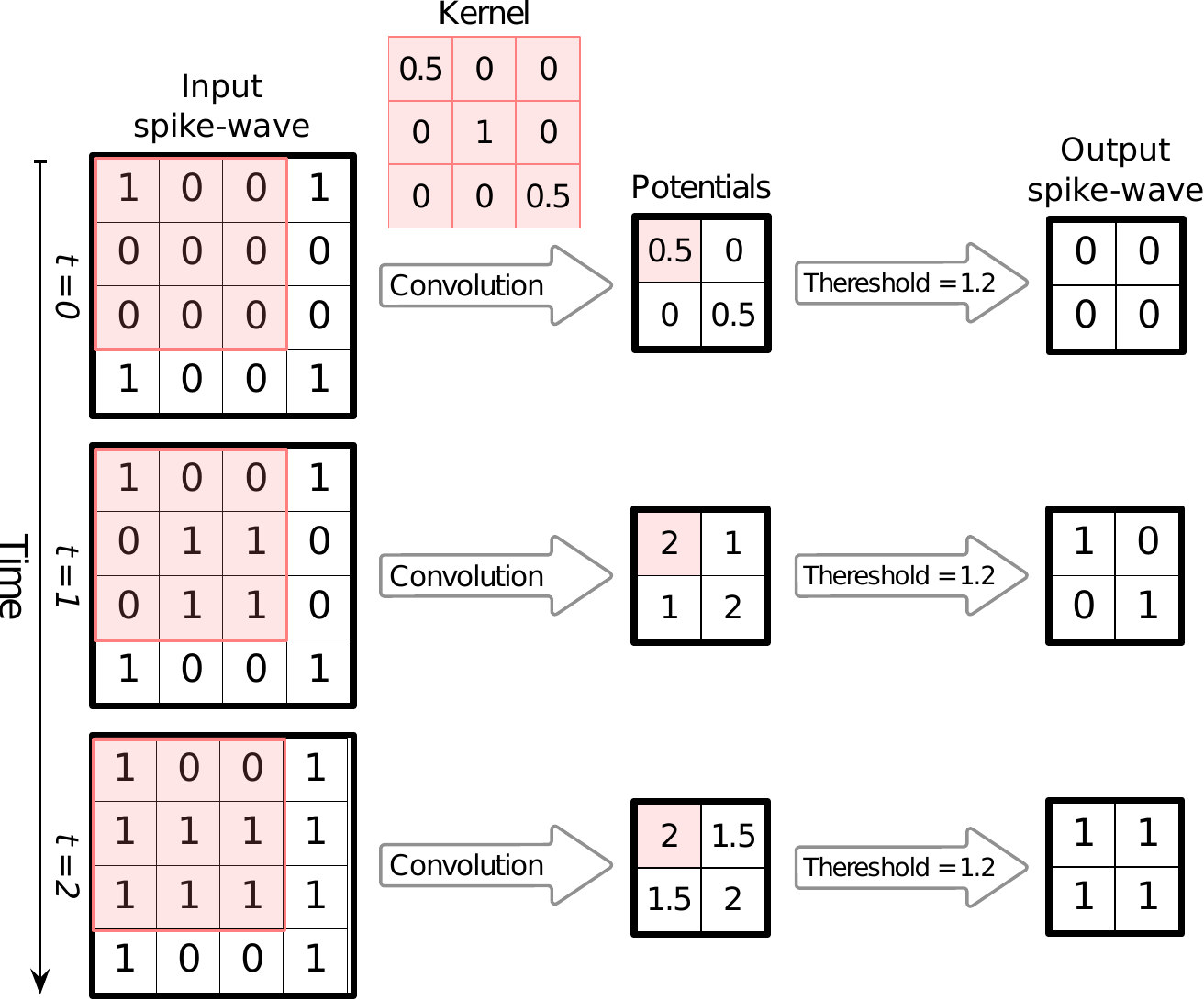}
	\end{center}
	\caption{An example of simultaneous processing of spikes over time-steps. Here the input spike-wave tensor has one $5\times5$ channel (feature map) and the spikes are divided into three time-steps. When SpykeTorch applies the convolution kernel of size $3\times 3$ (valid mode) simultaneously on all of the time-steps, the resulting tensor will contain potentials in all of the time-steps. Since spikes are stored in accumulative format, then the potentials are accumulative as well. Applying a threshold function over the whole potential tensor generates the corresponding output spike-wave tensor, again in accumulative format.}\label{fig:2}
\end{figure}

Note that this way of keeping the spikes (accumulative structure) does not mean that neurons keep firing after their first spikes. Repeating spikes in future time steps increases the memory usage, but makes it possible to process all of the time-steps simultaneously and produce the corresponding outputs, which consequently results in a huge speed-up. Figure \ref{fig:1} illustrates an example of converting spike times into a SpykeTorch-compatible spike-wave tensor. Figure \ref{fig:2} shows how accumulative spikes helps simultaneous computations.

\section{Package Structure}\label{pkg}

Basically, SpykeTorch consists of four python modules; (1) \texttt{snn} which contains multiple classes for creating SNNs, (2) \texttt{functional} that implements useful SNNs' functions, (3) \texttt{utils} which gathers helpful utilities, and (4) \texttt{visualization} which helps to generate graphical data out of SNNs. The following subsections explain these modules.

\subsection{\texttt{snn} Module}
The \texttt{snn} module contains necessary classes to build SNNs. These classes are inherited from the PyTorch's \texttt{nn.Module}, enabling them to function inside the PyTorch framework as network modules. Since we do not support error backpropagation, the PyTorch's auto-grad feature is turned off for all of the parameters in \texttt{snn} module.

\texttt{snn.Convolutional} objects implements spiking convolutional layers with two-dimensional convolution kernel. A \texttt{snn.Convolutional} object is built by providing the number of input and output features (or channels), and the size of the convolution kernel. Given the size of the kernel, the corresponding tensor of synaptic weights is randomly initialized using a normal distribution, where the mean and standard deviation can be set for each object, separately.

A \texttt{snn.Convolutional} object with kernel size $K_h\times K_w$ performs a valid convolution (with no padding) over an input spike-wave tensor of size $T_{max}\times F_{in}\times H_{in}\times W_{in}$ with stride equals to $1$ and produces an output potentials tensor of size $T_{max}\times F_{out}\times H_{out}\times W_{out}$, where:
\begin{equation}
\begin{split}
	H_{out} &= H_{in} - K_h + 1,\\
	W_{out} &= W_{in} - K_w + 1,
\end{split}
\end{equation}
and $F_{in}$ and $F_{out}$ are the number of input and output features, respectively. Potentials tensors ($P$) are similar to the binary spike-wave tensors, however $P[t,f,r,c]$ denotes the floating-point potential of a neuron placed at position $(r,c)$ of feature map $f$, at time-step $t$. Note that current version of SpykeTorch does not support stride more than $1$, however, we are going to implement it in the next major version.

The underlying computation of \texttt{snn.Convolutional} is the PyTorch's two-dimensional convolution, where the mini-batch dimension is sacrificed for the time. According to the accumulative structure of spike-wave tensor, the result of applying PyTorch's \texttt{conv2D} over this tensor is the accumulative potentials over time-steps.

It is important to mention that simultaneous computation over time dimension improves the efficiency of the framework, but it has dispelled batch processing in SpykeTorch. We agree that batch processing brings a huge speed-up, however, providing it to the current version of SpykeTorch is not straightforward. Here are some of the important challenges: (1) Due to accumulative format of spike-wave tensors, keeping batch of images increases memory usage even more. (2) Plasticity in batch mode needs new strategies. (3) To get the most out of batch processing, all of the main computations such as plasticity, competitions, and inhibitions should be done on the whole batch at the same time, especially when the model is running on GPUs.

Pooling is an important operation in deep convolutional networks. \texttt{snn.Pooling} implements two-dimensional max-pooling operation. Building \texttt{snn.Pooling} objects requires providing the pooling window size. The stride is equal to the window size by default, but it is adjustable. Zero padding is also another option which is off by default.

\texttt{snn.Pooling} objects are applicable to both spike-wave and potentials tensors. According to the structure of these tensors, if the input is a spike-wave tensor, then the output will contain the earliest spike within each pooling window, while if the input is a potentials tensor, the maximum potential within each pooling window will be extracted. Assume that the input tensor has the shape $T_{max}\times F_{in}\times H_{in}\times W_{in}$, the pooling window has the size $P_h\times P_w$ with stride $R_h\times R_w$, and the padding is $(D_{h}, D_{w})$, then the output tensor will have the size $T_{max}\times F_{out}\times H_{out}\times W_{out}$, where:
\begin{equation}
	\begin{split}
	H_{out} &= \lfloor \frac{H_{in} + 2\times D_{h}}{R_h} \rfloor,\\
	W_{out} &= \lfloor \frac{W_{in} + 2\times D_{w}}{R_w} \rfloor.
	\end{split}
\end{equation}

To apply STDP on a convolutional layer, a \texttt{snn.STDP} object should be built by providing the value of required parameters such as learning rates. Since this simulator works with time-to-first-spike coding, the provided implementation of the STDP function is as follows:

\begin{equation}
	\Delta \mathcal{W}_{i,j} = \begin{cases}
	A^+\times (\mathcal{W}_{i,j} - LB) \times (UB - \mathcal{W}_{i,j}) & \textit{if\ \ \ } T_j \leq T_i,\\
	A^-\times (\mathcal{W}_{i,j} - LB) \times (UB - \mathcal{W}_{i,j}) & \textit{if\ \ \ } T_j > T_i,
	\end{cases}
\end{equation}
where, $\Delta \mathcal{W}_{i,j}$ is the amount of weight change of the synapse connecting the post-synaptic neuron $i$ to the pre-synaptic neuron $j$, $A^+$ and $A^-$ are learning rates, and $(\mathcal{W}_{i,j} - LB) \times (UB - \mathcal{W}_{i,j})$ is a stabilizer term which slows down the weight change when the synaptic weight ($\mathcal{W}_{i,j}$) is close to the lower ($LB$) or upper ($UB$) bounds.

To apply STDP during the training process, providing the input and output spike-wave, as well as output potentials tensors are necessary. \texttt{snn.STDP} objects make use of the potentials tensor to find winners. Winners are selected first based on the earliest spike times, and then based on the maximum potentials. The number of winners is set to $1$ by default. \texttt{snn.STDP} objects also provide lateral inhibition, by which they completely inhibit the winners' surrounding neurons in all of the feature maps within a specific distance. This increases the chance of learning diverse features. Note that R-STDP can be applied using two \texttt{snn.STDP} objects; one for STDP part and the other for anti-STDP part.

\subsection{\texttt{functional} Module}
This module contains several useful and popular functions applicable on SNNs. Here we briefly review the most important ones. For the sake of simplicity, we replace the \texttt{functional.} with \texttt{sf.} for writing the function names.

As mentioned before, \texttt{snn.Convolutional} objects give potential tensors as their outputs. \texttt{sf.fire} takes a potentials tensor as input and converts it into a spike-wave tensor based on a given threshold. \texttt{sf.threshold} function is also available separately that takes a potentials tensor and outputs another potentials tensor in which all of the potentials lower than the given threshold are set to zero. The output of \texttt{sf.threshold} is called thresholded potentials.

Lateral inhibition is another vital operation for SNNs specially during the training process. It helps to learn more diverse features and achieve sparse representations in the network. SpykeTorch's \texttt{functional} module provides several functions for different kinds of lateral inhibitions.

\texttt{sf.feature\_inhibition} is useful for complete inhibition of the selected feature maps. This function comes in handy to apply dropout to a layer. \texttt{sf.pointwise\_inhibition} employs competition among features.  In other words, at each location, only the neuron corresponding to the most salient feature will be allowed to emit a spike (the earliest spike with the highest potential). Lateral inhibition is also helpful to be applied on input intensities before conversion to spike-waves. This will decrease the redundancy in each region of the input. To apply this kind of inhibition, \texttt{sf.intensity\_lateral\_inhibition} is provided. It takes intensities and a lateral inhibition kernel by which it decreases the surrounding intensities (thus increases the latency of the corresponding spike) of each salient point. Local normalization is also provided by \texttt{sf.local\_normalization} which uses regional mean for normalizing intensity values.

Winners-take-all (WTA) is a popular competition mechanism in SNNs. WTA is usually used for plasticity, however, it can be involved in other functionalities such as decision-making. \texttt{sf.get\_k\_winners} takes the desired number of winners and the thresholded potentials and returns the list of winners. Winners are selected first based on the earliest spike times, and then based on the maximum potentials. Each winner's location is represented by a triplet of the form $(feature, row, column)$.

\subsection{\texttt{utils} Module}
\texttt{utils} module provides several utilities to ease the implementation of ideas with SpykeTorch. For example, \texttt{utils.generate\_inhibition\_kernel} generates an inhibition kernel based on a series of inhibition factors in a form that can be properly used by \texttt{sf.intensity\_lateral\_inhibition}.

There exist several transformation utilities that are suitable for filtering inputs and converting them to spike-waves. Current utilities are mostly designed for vision purposes. \texttt{utils.LateralIntencityInhibition} objects do the \texttt{sf.intensity\_lateral\_inhibition} as a transform object. \texttt{utils.FilterKernel} is a base class to define filter kernel generators. SpykeTorch has already provided \texttt{utils.DoGKernel} and \texttt{utils.GaborKernel} in order to generate DoG and Gabor filter kernels, respectively. Objects of \texttt{utils.FilterKernel} can be packed into a multi-channel filter kernel by \texttt{utils.Filter} objects and applied to the inputs.

The most important utility provided by \texttt{utils} is \texttt{utils.Intensity2Latency}. Objects of \texttt{utils.Intensity2Latency} are used as transforms in PyTorch's datasets to transform intensities into latencies, i.e. spike-wave tensors. Usually, \texttt{utils.Intensity2Latency} is the final transform applied to inputs.

Since the application of a series of transformations and the conversion to spike-waves can be time-consuming, SpykeTorch provides a wrapper class, called \texttt{utils.CacheDataset}, which is  inherited from PyTorch's \texttt{dataset} class. Objects of \texttt{utils.CacheDataset} take a dataset as their input and cache the data after applying all of the transformations. They can cache either on primary memory or secondary storage.

Additionally, \texttt{utils} contains two functions \texttt{utils.tensor\_to\_text} and \texttt{utils.text\_to\_tensor}, which handle conversion of tensors to text files and the reverse, respectively. This conversion is helpful to import data from a source or export a tensor for a target software. The format of the text file is as follows: the first line contains comma-separated integers denoting the shape of the tensor. The second line contains comma-separated values indicating the whole tensor's data in row-major order.

\subsection{\texttt{visualization} Module}
The ability to visualize deep networks is of great importance since it gives a better understanding of how the network's components are working. However, visualization is not a straightforward procedure and depends highly on the target problem and the input data.

Due to the fact that SpykeTorch is developed mainly for vision tasks, its \texttt{visualization} module contains useful functions to reconstruct the learned visual features. The user should note that these functions are not perfect and cannot be used in every situation. In general, we recommend the user to define his/her own visualization functions to get the most out of the data.

\section{Tutorial}\label{tutor}
In this section, we show how to design, build, train, and test a SNN with SpykeTorch in a tutorial format. The network in this tutorial is adopted from the deep convolutional SNN proposed by~\cite{mozafari2019bioinspired} which recognizes handwritten digits (tested on MNIST dataset). This network has a deep structure and uses both STDP and Reward-Modulated STDP (R-STDP), which makes it a suitable choice for a complete tutorial. In order to make the tutorial as simple as possible, we present code snippets with reduced contents. For the complete source code, please check SpykeTorch's GitHub\footnote{\url{https://github.com/miladmozafari/SpykeTorch}} web page.

\subsection{Step 1. Network Design}
\subsubsection{Structure}
\begin{figure}
	\begin{center}
		\includegraphics[width=0.8\textwidth]{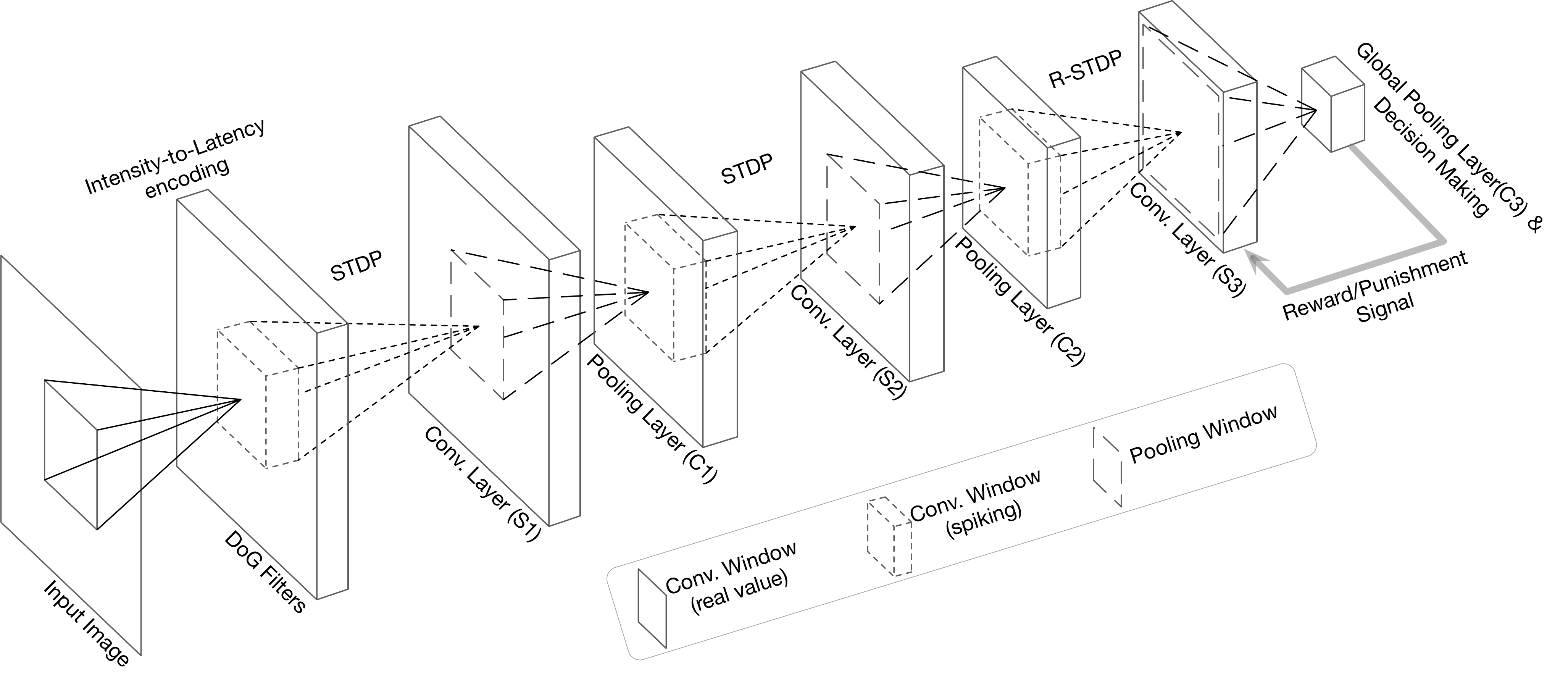}
	\end{center}
	\caption{Overall structure of the network in the tutorial (modified figure from~\cite{mozafari2019bioinspired}).}\label{fig:3}
\end{figure}

The best way to design a SNN is to define a class inherited from \texttt{torch.nn.Module}. The network proposed in~\cite{mozafari2019bioinspired}, has an input layer which applies DoG filters to the input image and converts it to spike-wave. After that, there are three convolutional ($S$) and pooling ($C$) layers that are arranged in the form of $S1 \rightarrow C1 \rightarrow S2 \rightarrow C2 \rightarrow S3 \rightarrow C3$ (see Figure \ref{fig:3}). Therefore, we need to consider three objects for convolutional layers in this model. For the pooling layers, we will use the functional version instead of the objects.

As shown in Listing \ref{lst:1}, three \texttt{snn.Convolutional} objects are created with desired parameters. Two \texttt{snn.STDP} objects are built for training $S1$ and $S2$ layers. Since $S3$ is trained by R-STDP, two \texttt{snn.STDP} are needed to cover both STDP and anti-STDP parts. To have the effect of anti-STDP, it is enough to negate the signs of the learning rates. Note that the \texttt{snn.STDP} objects for \texttt{conv3} have two extra parameters where the first one turns off the stabilizer and the second one keeps the weights in range $[0.2, 0.8]$.

Although \texttt{snn} objects are compatible with \texttt{nn.Sequential} (\texttt{nn.Sequential} automates the forward pass given the network modules), we cannot use it at the moment. The reason is that different learning rules may need different kinds of data from each layer, thus accessing each layer during the forward pass is a must.

\begin{code}[t]
	\centering
	\includegraphics{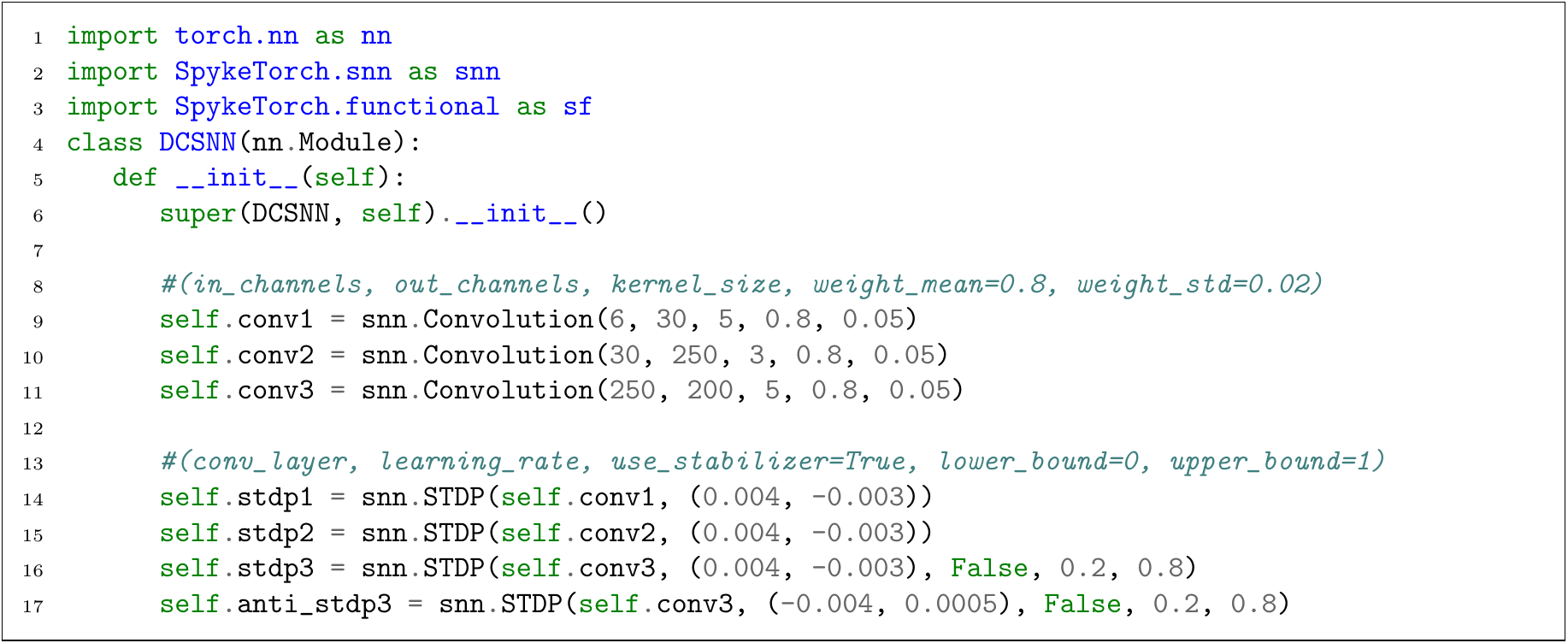}
	\caption{Defining the network class.}
	\label{lst:1}
\end{code}

\subsubsection{Forward Pass}
Next, we implement the forward pass of the network. To this end, we override the \texttt{forward} function in \texttt{nn.Module}. If the training is off, then implementing the forward pass will be straightforward. Listing \ref{lst:2} shows the application of convolutional and pooling layers on an input sample. Note that each input is a spike-wave tensor. We will demonstrate how to convert images into spike-waves later.

\begin{code}[t]
	\centering
	\includegraphics{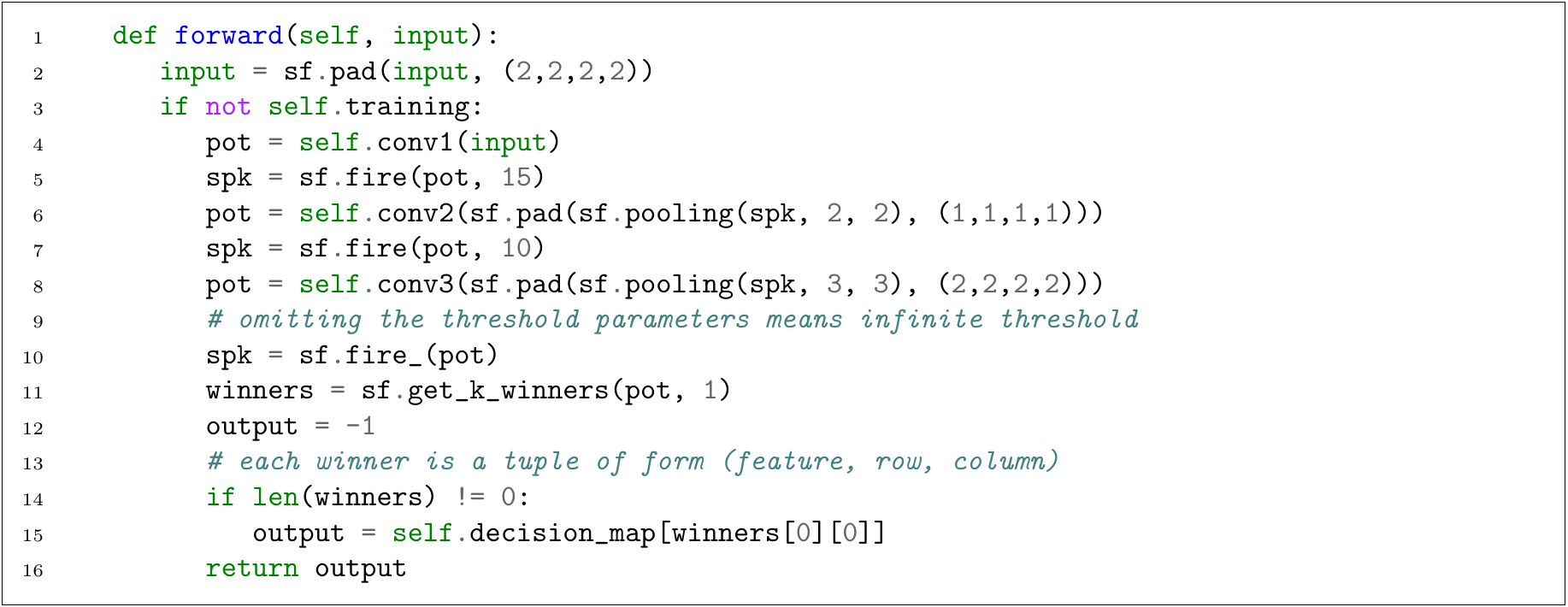}
	\caption{Defining the forward pass (during testing process).}
	\label{lst:2}
\end{code}

As shown in Listing \ref{lst:2}, the input of each convolutional layer is the padded version of the output of its previous layer, thus, there would be no information loss at the boundaries. Pooling operations are also applied by the corresponding function \texttt{sf.pooling}, which is an alternative to \texttt{snn.Pooling}. According to~\cite{mozafari2019bioinspired}, their proposed network makes decision based on the maximum potential among the neurons in the last pooling layer. To this end, we use an infinite threshold for the last convolutional layer by omitting its value from \texttt{sf.fire\_} function. \texttt{sf.fire\_} is the in-place version of \texttt{sf.fire} which modifies the input potentials tensor $P_{in}$ as follows:
\begin{equation}
	P_{in}[t,f,r,c] = \begin{cases}
	0 & \textit{if\ \ \ } t < T_{max} - 1,\\
	P_{in}[t,f,r,c] & \textit{otherwise.}
	\end{cases}
\end{equation}
Consequently, the resulting spike-wave will be a tensor in which, all the values are zero except for those non-zero potential values in the last time-step.

Now that we have the potentials of all the neurons in $S3$, we find the only one winner among them. This is the same as doing a global max-pooling and choosing the maximum potential among them. \texttt{decision\_map} is a Python list which maps each feature to a class label. Since each winner contains the feature number as its first component, we can easily indicate the decision of the network by putting that into the \texttt{decision\_map}.

We cannot take advantage of this forward pass during the training process as the STDP and R-STDP need local synaptic data to operate. Therefore, we need to save the required data during the forward pass. We define a Python dictionary (named \texttt{ctx}) in our network class and a function which saves the data into that (see Listing \ref{lst:3}). Since the training process is layer-wise, we update the \texttt{forward} function to take another parameter which specifies the layer that is under training. The updated \texttt{forward} function is shown in Listing \ref{lst:4}.

\begin{code}[t]
	\centering
	\includegraphics{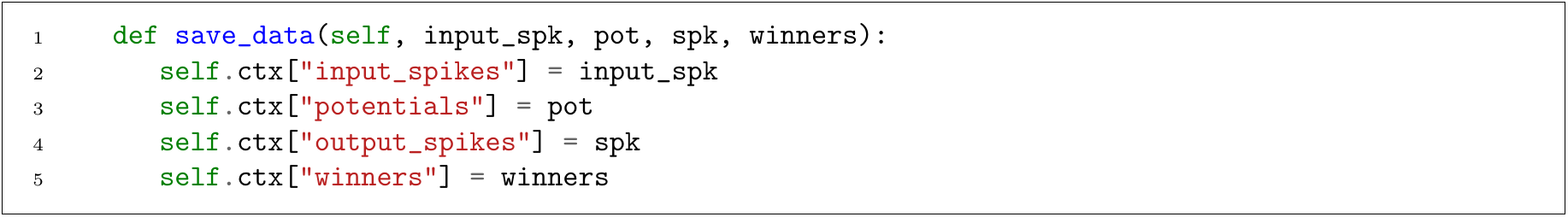}
	\caption{Saving required data for plasticity.}
	\label{lst:3}
\end{code}

\begin{code}[t]
	\centering
	\includegraphics{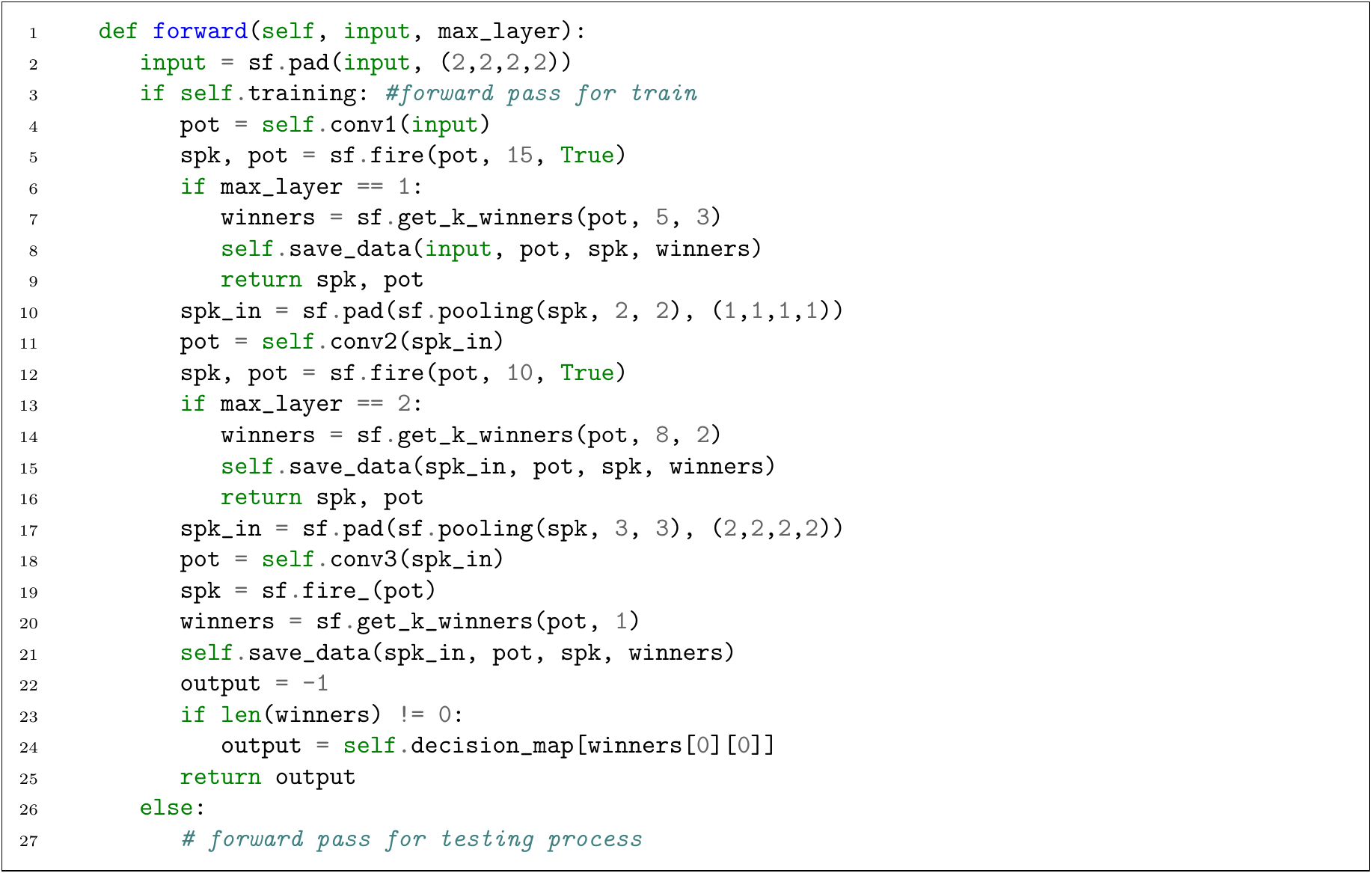}
	\caption{Defining the forward pass (during training process).}
	\label{lst:4}
\end{code}

There are several differences with respect to the testing forward pass. First, \texttt{sf.fire} is used with an extra parameter value. If the value of this parameter is \texttt{True}, the tensor of thresholded potentials will also be returned. Second, \texttt{sf.get\_k\_winners} is called with a new parameter value which controls the radius of lateral inhibition. Third, the forward pass is interrupted by the value of \texttt{max\_layer}.

\subsubsection{Plasticity}
Now that we saved the required data for learning, we can define a series of helper functions to apply STDP or anti-STDP. Listing \ref{lst:5} defines three member functions for this purpose. For each call of \texttt{STDP} objects, we need to provide tensors of input spike-wave, output thresholded potentials, output spike-wave, and the list of winners.

\begin{code}[t]
	\centering
	\includegraphics{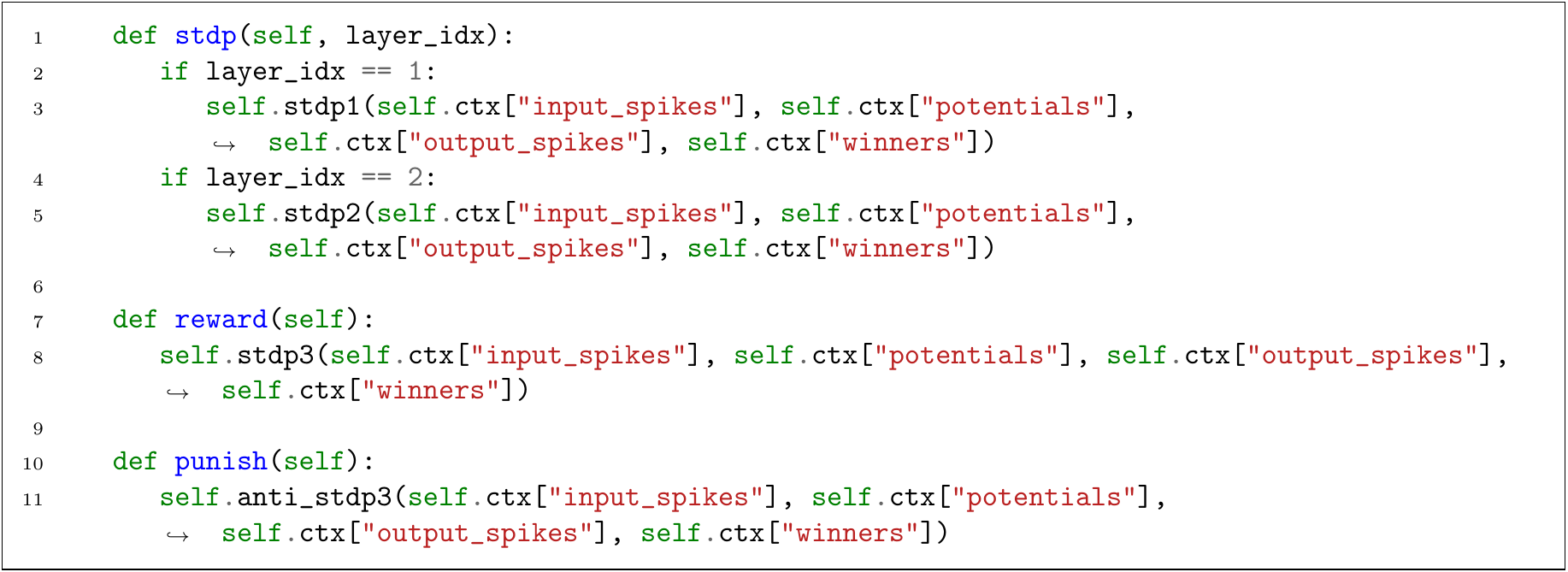}
	\caption{Defining helper functions for plasticity.}
	\label{lst:5}
\end{code}

\subsection{Step 2. Input Layer and Transformations}
SNNs work with spikes, thus, we need to transform images into spike-waves before feeding them into the network. PyTorch's datasets accept a function as a transformation which is called automatically on each input sample. We make use of this feature together with the provided transform functions and objects by PyTorch and SpykeTorch. According to the network proposed in~\cite{mozafari2019bioinspired}, each image is convolved by six DoG filters, locally normalized, and transformed into spike-wave. As appeared in Listing \ref{lst:6}, a new class is defined to handle the required transformations.

\begin{code}[t]
	\centering
	\includegraphics{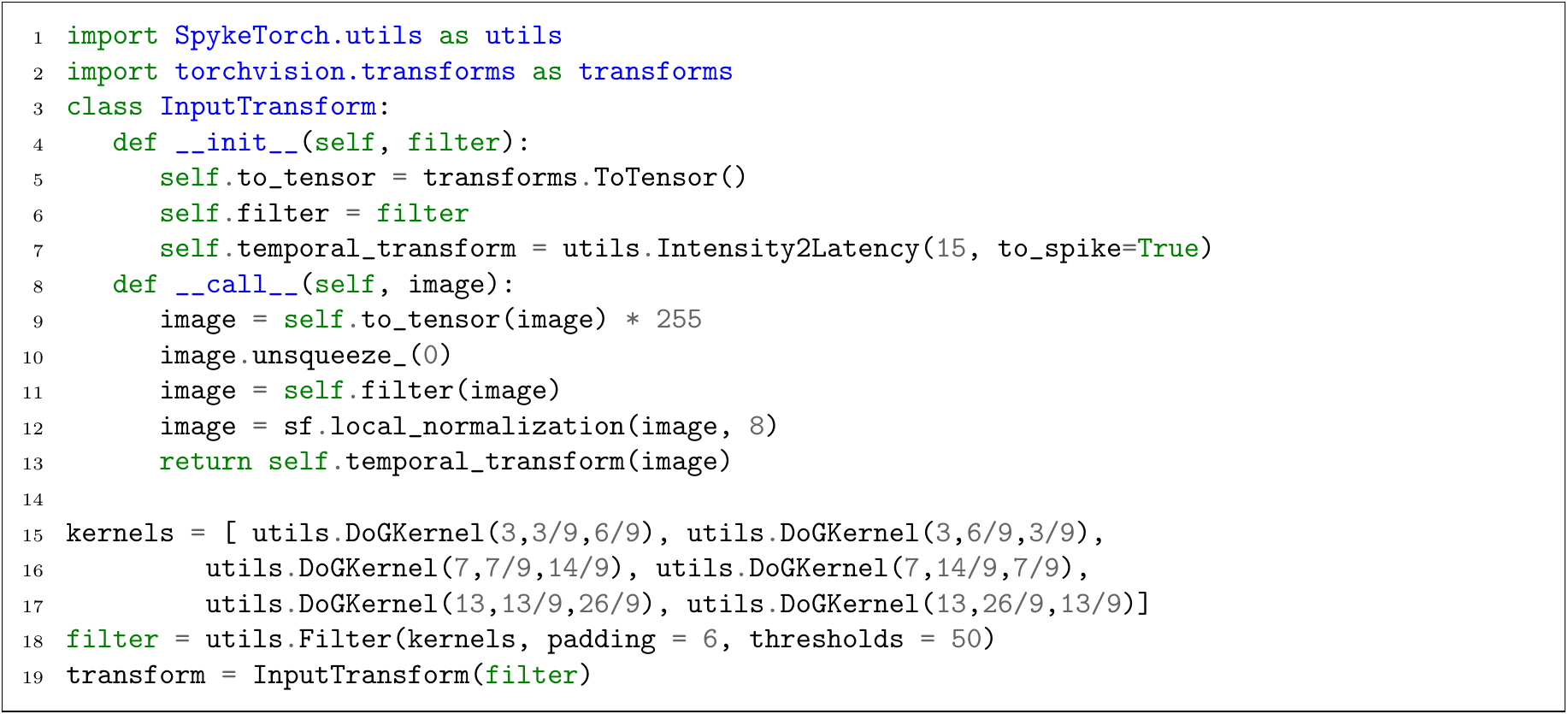}
	\caption{Transforming each input image into spike-wave.}
	\label{lst:6}
\end{code}

Each \texttt{InputTransform} object converts the input image into a tensor (line 9), adds an extra dimension for time (line 10), applies provided filters (line 11), applies local normalization (line 12), and generates spike-wave tensor (line 13). To create \texttt{utils.Filter} object, six DoG kernels with desired parameters are given to \texttt{utils.Filter}'s constructor (lines 15-17) as well as an appropriate padding and threshold value (line 18).

\subsection{Step 3. Data Preparation}
Due to the PyTorch and SpykeTorch compatibility, all of the PyTorch's dataset utilities work here. As illustrated in Listing \ref{lst:7}, we use \texttt{torchvision.datasets.MNIST} to load MNIST dataset with our previously defined \texttt{transform}. Moreover, we use SpykeTorch's dataset wrapper, \texttt{utils.CacheDataset} to enable caching the transformed data after its first presentation. When the dataset gets ready, we use PyTorch's \texttt{DataLoader} to manage data loading.
\newline
\begin{code}[t]
	\centering
	\includegraphics{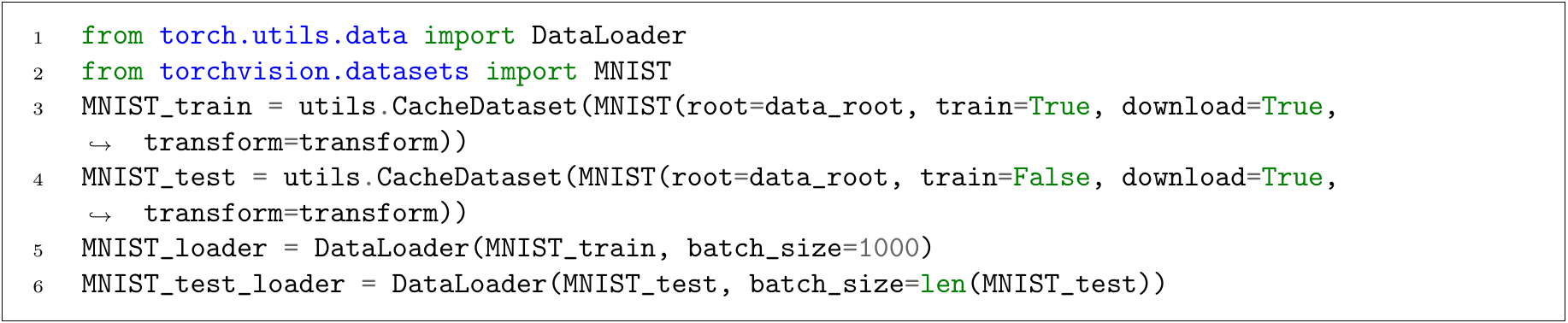}
	\caption{Preparing MNIST dataset and the data loader.}
	\label{lst:7}
\end{code}

\subsection{Step 4. Training and Testing}

\subsubsection{Unsupervised Learning (STDP)}
To do unsupervised learning on $S1$ and $S2$ layers, we use a helper function as defined in Listing \ref{lst:8}. This function trains layer \texttt{layer\_idx} of \texttt{network} on \texttt{data} by calling the corresponding \texttt{STDP} object. There are two important things in this function: (1) putting the network in train mode by calling \texttt{.train} function, and (2) loading the sample on GPU if the global \texttt{use\_cuda} flag is \texttt{True}.
\newline

\begin{code}[t]
	\centering
	\includegraphics{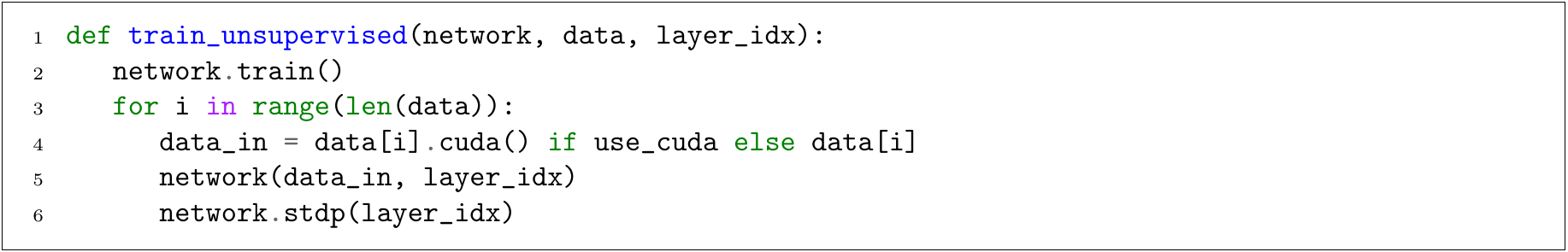}
	\caption{Helper function for unsupervised learning.}
	\label{lst:8}
\end{code}

\subsubsection{Reinforcement Learning (R-STDP)}
To apply R-STDP, it is enough to call previously defined \texttt{reward} or \texttt{punish} member functions under appropriate conditions. As shown in Listing \ref{lst:9}, we check the network's decision with the label and call \texttt{reward} (or \texttt{punish}) if it matches (or mismatches) the target. We also compute the performance by counting correct, wrong, and silent (no decision is made because of receiving no spikes) samples.
\newline
\begin{code}[t]
	\centering
	\includegraphics{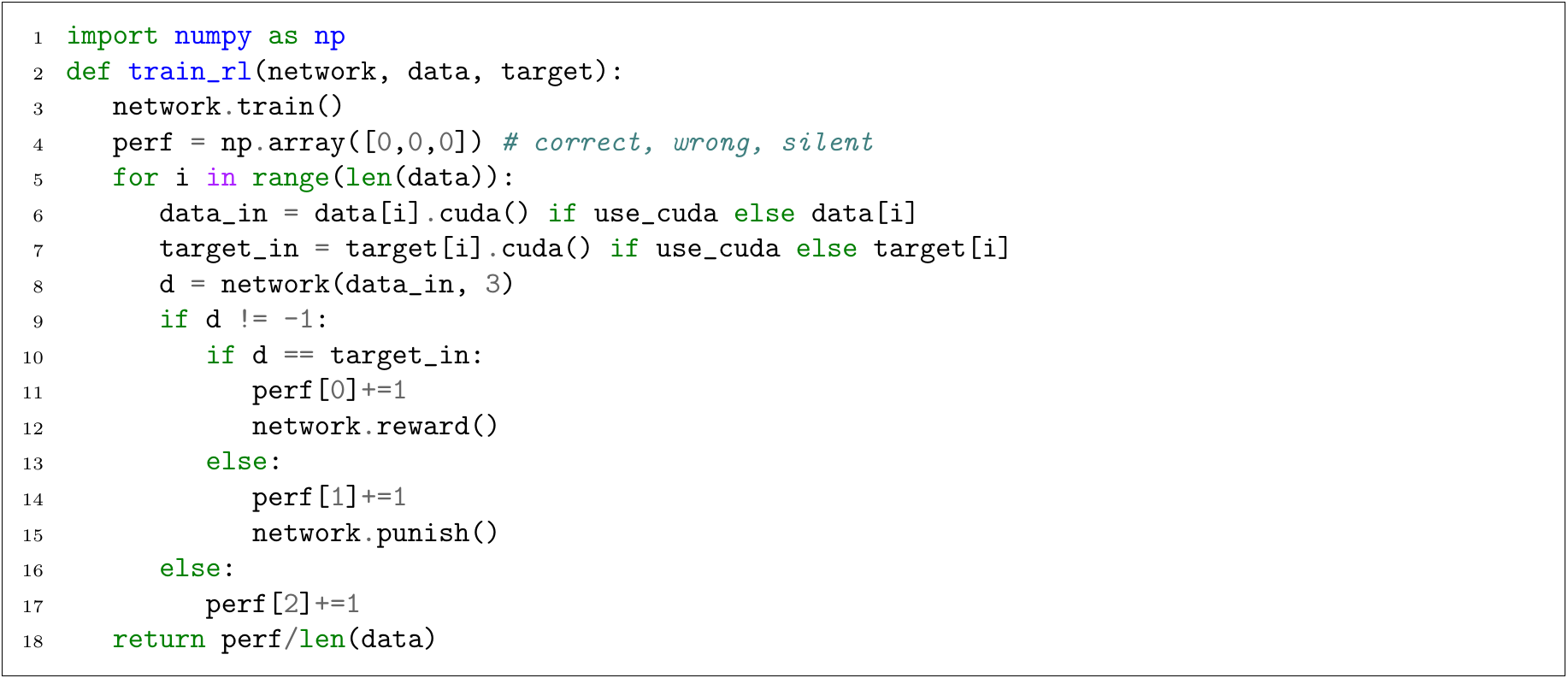}
	\caption{Helper function for reinforcement learning.}
	\label{lst:9}
\end{code}

\subsubsection{Execution}
Now that we have the helper functions, we can make an instance of the network and start training and testing it. Listing \ref{lst:10} illustrates the implementation of this part. Note that the \texttt{test} helper function is the same as the \texttt{train\_rl} function, but it calls \texttt{network.eval} instead of \texttt{network.train} and it does not call plasticity member functions. Also, invoking \texttt{net.cuda}, transfers all the network's parameters to the GPU.
\newline
\begin{code}[t]
	\centering
	\includegraphics{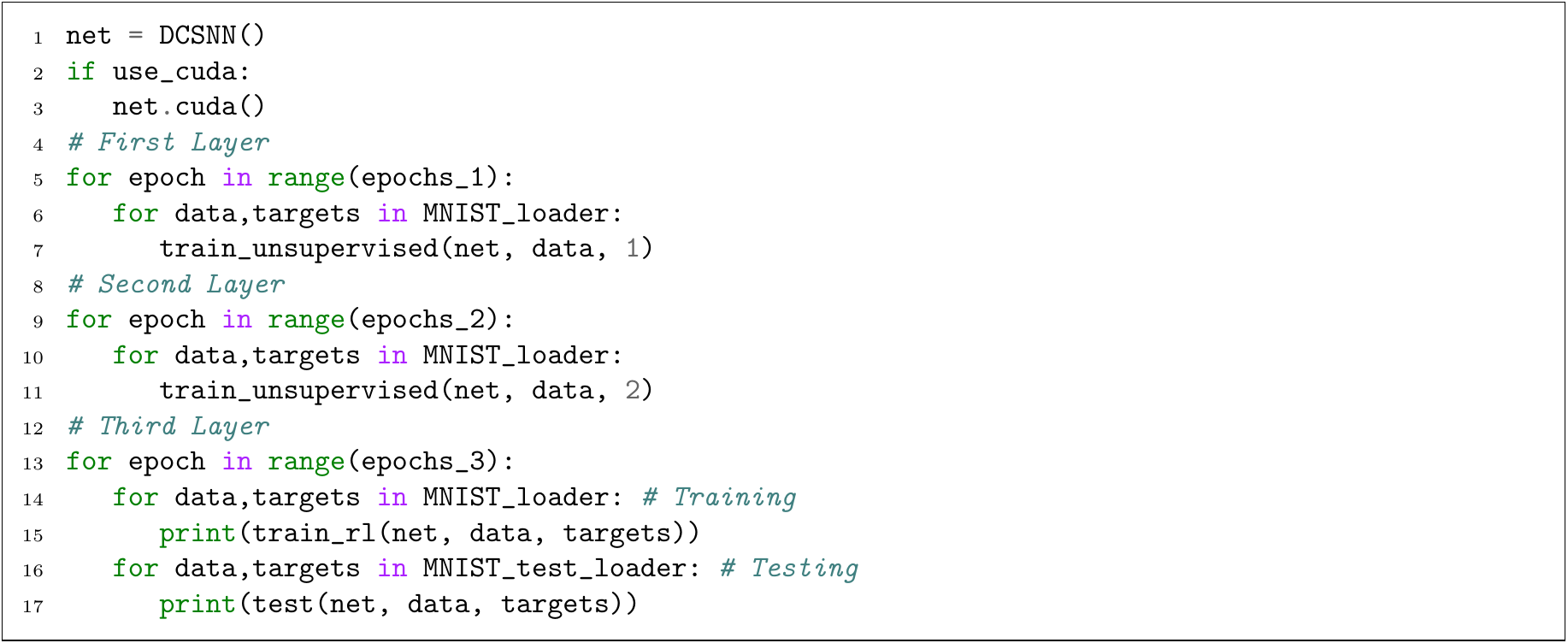}
	\caption{Training and testing the network.}
	\label{lst:10}
\end{code}

\subsection{Source Code}
Through this tutorial, we omitted many parts of the actual implementation such as adaptive learning rates, multiplication of learning rates, and saving/loading the best state of the network, for the sake of simplicity and clearance. The complete reimplementation is available on SpykeTorch's GitHub web page. We have also provided scripts for other works~\cite{mozafari2018first, kheradpisheh2018stdp} that achieve almost the same results as the main implementations. However, due to technical and computational differences between SpykeTorch and the original versions, tiny differences in performance are expected. A comparison between SpykeTorch and one of the previous implementations is provided in the next section.


\section{Comparison}\label{compare}
We performed a comparison between SpykeTorch and the dedicated C++/CUDA implementations of the network proposed in~\cite{mozafari2019bioinspired} and measured the training and inference time. Both networks are trained for $686$ epochs ($2$ for the first, $4$ for the second, and $680$ for the last trainable layer). In each training or inference epoch, the network sees all of the training or testing samples respectively. Note that during the training of the last trainable layer, each training epoch is followed by an inference epoch.

As shown in Table~\ref{tbl:1}, SpykeTorch script outperformed the original implementation in both training and inference times. The small performance gap is due to some technical differences in functions' implementations and performing a new round of parameter tuning fills this gap. We believe that SpykeTorch has the potentials of even more efficient computations. For example, adding batch processing to SpykeTorch would result in a large amount of speed-up due to the minimization of CPU-GPU interactions.

\renewcommand{\arraystretch}{1.5}
\begin{table}
	\centering
	\caption{Comparison of C++/CUDA and SpykeTorch scripts simulating the network proposed in~\cite{mozafari2019bioinspired}. Both scripts are executed on a same machine with Intel(R) Xeon(R) CPU E5-2697 (2.70 GHz), 256G Memory, NVIDIA TITAN Xp GPU, PyTorch 1.1.0, and Ubuntu 16.04.}
	\label{tbl:1}
	\begin{tabular}{|c|c|c|c|}
		\hline
		Script & Total Training & Inference Epoch & Accuracy\\
		\hline
		\hline
		C++/CUDA & $174120$s ($= 2\text{d }00\text{h }22\text{m }20\text{s}$) & $35$s & $97.2\%$ \\
		\hline
		SpykeTorch & $121600$s ($= 1\text{d }09\text{h }46\text{m }40\text{s}$) & $20$s & $96.9\%$ \\
		\hline
	\end{tabular}
\end{table}

\section{Conclusions}\label{conclusion}
In recent years, SNNs have gained many interests in AI because of their ability to work in a spatio-temporal domain as well as energy efficiency. Unlike DCNNs, most of the current SNN simulators are not efficient enough to perform large-scale AI tasks. In this paper, we proposed SpykeTorch, an open-source high-speed simulation framework based on PyTorch. The proposed framework is optimized for convolutional SNNs with at most one spike per neuron and time-to-first-spike information coding scheme. SpykeTorch provides STDP and R-STDP learning rules but other rules can be added easily.

The compatibility and integrity of SpykeTorch with PyTorch have simplified its usage specially for the deep learning communities. This integration brings almost all of the PyTorch's features functionalities to SpykeTorch such as the ability of just-in-time optimization for running on CPUs, GPUs, or Multi-GPU platforms. We agree that SpykeTorch has hard limitations on type of SNNs, however, there is always a trade-off between computational efficiency and generalization. Apart from the increase of computational efficiency, this particular type of SNNs are bio-realistic, energy-efficient, and hardware-friendly that are getting more and more popular recently.

We provided a tutorial on how to build, train, and evaluate a DCSNN for digit recognition using SpykeTorch. However, the resources are not limited to this paper and additional scripts and documentations can be found on SpykeTorch's GitHub page. We reimplemented various works~\cite{mozafari2018first,kheradpisheh2018stdp, mozafari2019bioinspired} by SpykeTorch and reproduced their results with negligible difference.

Although the current version of SpykeTorch is functional and provides the main modules and utilities for DCSNNs (with at most one spike per neuron), we will not stop here and our plan is to extend and improve it gradually. For example, adding automation utilities would ease programming the network's forward pass resulting a more readable and cleaner code. Due to the variations of training strategies, designing a general automation platform is challenging. Another feature that improves SpykeTorch's speed is batch processing. Enabling batch mode might be easy for operations like convolution or pooling, however, implementing batch learning algorithms that can be run with none or a few CPU-GPU interactions is hard. Finally, implementing features to support models for other modalities such as the auditory system makes SpykeTorch a multi-modal SNN framework.

\section*{Acknowledgments}
This research was partially supported by the Iranian Cognitive Sciences and Technologies Council (Grant no. 5898) and by the French Agence Nationale de la Recherche (grant: Beating Roger Federer ANR-16-CE28-0017-01).

The authors would like to thank Dr. Jean-Pierre Jaffr\'ezou
for proofreading this manuscript and NVIDIA GPU Grant Program for supporting computations by providing a high-tech GPU.

\footnotesize
\bibliographystyle{elsarticle-num}

\end{document}